\documentclass[letterpaper]{article} 
\usepackage{aaai2026}  
\usepackage{times}  
\usepackage{helvet}  
\usepackage{courier}  
\usepackage[hyphens]{url}  
\usepackage{graphicx} 
\usepackage{natbib}  
\usepackage{caption} 
\usepackage{algorithm}
\usepackage{algorithmic}
\usepackage{amssymb}
\usepackage{amsmath}
\usepackage{multirow}
\usepackage{multicol}
\usepackage{newfloat}
\usepackage{listings}
\DeclareCaptionStyle{ruled}{labelfont=normalfont,labelsep=colon,strut=off} 
\floatstyle{ruled}
\newfloat{listing}{tb}{lst}{}
\floatname{listing}{Listing}
\title{MirrorSAM2: Segment Mirror in Videos with Depth Perception}

\author{\textbf{Mingchen Xu}\textsuperscript{1} 
\quad \textbf{Yu-Kun Lai}\textsuperscript{1} 
\quad \textbf{Ze Ji}\textsuperscript{2} 
\quad \textbf{Jing Wu}\textsuperscript{1}\\
\textsuperscript{1}School of Computer Science and Informatics, Cardiff University\\
\textsuperscript{2}School of Engineering, Cardiff University\\
{\tt\small \{xum35, laiy4, jiz1, wuj11\}@cardiff.ac.uk}
}
\usepackage{bibentry}

\begin{document}

\maketitle

\begin{abstract}
This paper presents MirrorSAM2, the first framework that adapts Segment Anything Model 2 (SAM2) to the task of RGB-D video mirror segmentation. MirrorSAM2 addresses key challenges in mirror detection—such as reflection ambiguity and texture confusion—by introducing four tailored modules: a Depth Warping Module for RGB-depth alignment, a Depth-guided Multi-Scale Point Prompt Generator for automatic prompt generation, a Frequency Detail Attention Fusion Module to enhance structural boundaries, and a Mirror Mask Decoder with a learnable mirror token for refined segmentation. By fully leveraging the complementarity between RGB and depth, MirrorSAM2 extends SAM2’s capabilities to the prompt-free setting. To our knowledge, this is the first work to enable SAM2 for automatic video mirror segmentation. Experiments on the VMD and DVMD-D benchmark demonstrate that MirrorSAM2 achieves state-of-the-art performance, even under challenging conditions such as small mirrors, weak boundaries, and strong reflections.
\end{abstract}


\section{Introduction}

\begin{figure}[t]
\centering
\includegraphics[width=1\columnwidth]{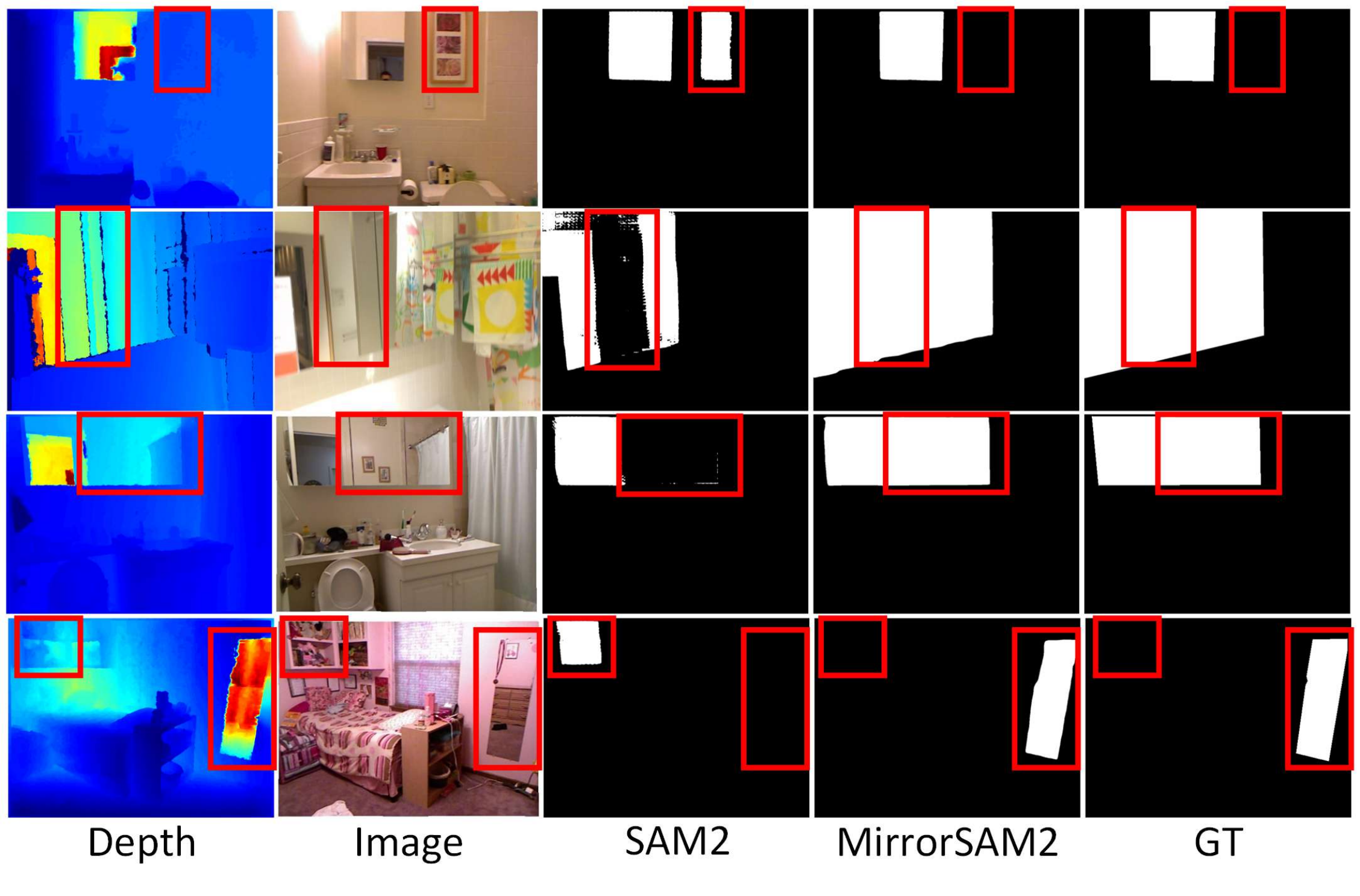} 
\caption{Visual comparison between the results of our proposed MirrorSAM2 and the original SAM2 \cite{ravi2024sam}. 
The $1^{st}$ column shows that depth maps align well with the mirror regions. The $1^{st}$ and $4^{th}$ rows reflect the role of the Depth Warping Module and Depth-guided Multi-scale Point Prompt Generator, which effectively suppresses incorrect segmentation by distinguishing between mirrors and visually similar objects with the help of depth features. The $2^{nd}$, $3^{rd}$, and $4^{th}$ rows embody the role of the Frequency Detail Attention Fusion Module and the Mirror Mask Decoder. They help recover previously missed mirror regions by enhancing fine-grained details and generating explicit mirror-aware predictions.}
\label{intro1}
\end{figure}

Segment Anything Model 2 (SAM2) \cite{ravi2024sam} is a recently introduced vision foundation model that has significantly advanced video segmentation. Trained on the large-scale SA-1B \cite{kirillov2023segment} (11M images, 1B masks) and SA-V \cite{ravi2024sam} (50.9K videos, 35.5M masks) datasets, SAM2 learns highly generalizable representations for natural scenes. However, its performance on the video mirror detection (VMD) task remains underexplored. While SAM2 performs well when provided with accurate mask prompts, its performance degrades significantly under point prompts, often failing to delineate complete mirror regions, as reported in \cite{jie2024sam2meetsvideoshadow}. Moreover, to the best of our knowledge, there has been no investigation into SAM2’s capabilities under prompt-free conditions. In this work, we present the first attempt to adapt SAM2 for RGB-D video mirror detection, in contrast to prior semi-supervised approaches that rely on manually provided or interaction-based prompts. By addressing the limitations of SAM2 in handling mirror ambiguity and reflection confusion without human intervention, our method enables robust and automatic mirror segmentation in videos. 

To address the challenging task of RGB-D video mirror detection, we propose a SAM2-based RGB-D segmentation framework named MirrorSAM2 (Segment Anything Model 2 with Depth Perception). MirrorSAM2 extends the strong segmentation capability of SAM2 by incorporating depth perception and structural priors, enabling accurate mirror segmentation without relying on manual prompts during inference. Mirrors are inherently visually deceptive—they reflect surrounding content and often share similar textures and boundaries with non-mirror regions in the RGB domain, making them difficult to distinguish. In contrast, depth maps reveal clear structural disparities between mirrors and their environment. MirrorSAM2 leverages this RGB-D complementarity to compensate for information loss in RGB features and improve segmentation robustness in such ambiguous scenarios. Specifically, MirrorSAM2 introduces four core modules: (1) The Depth Warping (DW) Module establishes cross-modal alignment between RGB and depth via bidirectional correlation and PAC-guided decoding, reducing semantic noise. (2) The Depth-guided Multi-scale Point Prompt Generator (DMS-PPG) extracts hierarchical structural cues to generate confident spatial prompts without supervision. (3) The Frequency Detail Attention Fusion (FDAF) Module transforms depth features into the frequency domain and aligns amplitude-phase information to recover fine-grained structure.(4) The Mirror Mask Decoder (MMD) integrates enhanced features with a learnable mirror token to explicitly guide mask prediction. Together, these modules allow MirrorSAM2 to accurately locate and segment mirrors in videos without any prompts. Compared to the original SAM2, our model effectively suppresses false positives and captures complete mirror regions. As illustrated in Fig.~\ref{intro1}, the DW and DMS-PPG modules are critical in correcting misclassifications caused by RGB-only ambiguity, while the FDAF and MMD modules improve boundary refinement and segmentation quality in challenging scenes.

Furthermore, there are noticeable omissions in the SAM2 segmentation results, as shown in the $2^{nd}$, $3^{rd}$, and $4^{th}$ of Fig. \ref{intro1}, where large portions of mirror regions are missing. To address this issue, we propose two key modules: the FDAF module and the MMD. The FDFA module operates on the early-stage features from the DW module, transforming them into the frequency domain to capture detailed structural information. By aligning amplitude and phase representations across scales, the FDAF module enables each memory token in the decoder to selectively attend to frequency components that are most relevant to its context. This frequency-guided attention mechanism enriches the model’s perception of subtle boundaries and fine-grained textures that are easily overlooked in the original SAM2 pipeline. Building on this, the MMD introduces a learnable mirror token that explicitly guides the model to focus on mirror-relevant semantics. It integrates both frequency-enhanced features and global contextual embeddings to produce more complete and accurate mirror masks. Together, these modules substantially reduce omission errors and improve the model’s ability to detect challenging mirror regions with weak or ambiguous boundaries. 

Our contributions are summarized as follows:
\begin{itemize}
    \item We propose MirrorSAM2, a variant of SAM2 enhanced with depth information and specifically tailored for the video mirror detection (VMD) domain. To the best of our knowledge, MirrorSAM2 is the first RGB-D model built upon the SAM2 framework for VMD, and also the first to adapt SAM2 to address this task in an automatic way. The model systematically explores the interaction between RGB and depth modalities within the SAM2 architecture, where depth serves as a complementary cue to RGB, enhancing semantic understanding and structural reasoning in challenging mirror segmentation scenarios. 
    \item Four modules are introduced in MirrorSAM2: the DW module, the DMS-PPG, the FDAF module, and the MMD. DW module enhances cross-modal alignment by establishing structural consistency between RGB and depth features, providing a reliable foundation for downstream processing. Based on this, DMS-PPG generates accurate and diverse point prompts by mining informative spatial regions from the depth-enhanced feature maps. FDAFM further refines structural perception by guiding memory tokens to focus on discriminative frequency-based cues, enabling the recovery of subtle mirror boundaries. Finally, MMD integrates these enhanced features with a learnable mirror token, producing explicit and complete mirror segmentation masks without any external prompts.
    \item Our approach achieves superior performance over 15 SOTA methods on the VMD and DVMD benchmark, demonstrating the effectiveness of RGB-D VMD segmentation under the SAM2 framework.

\end{itemize}

\begin{figure*}[ht]
\centering
\includegraphics[width=0.9\textwidth]{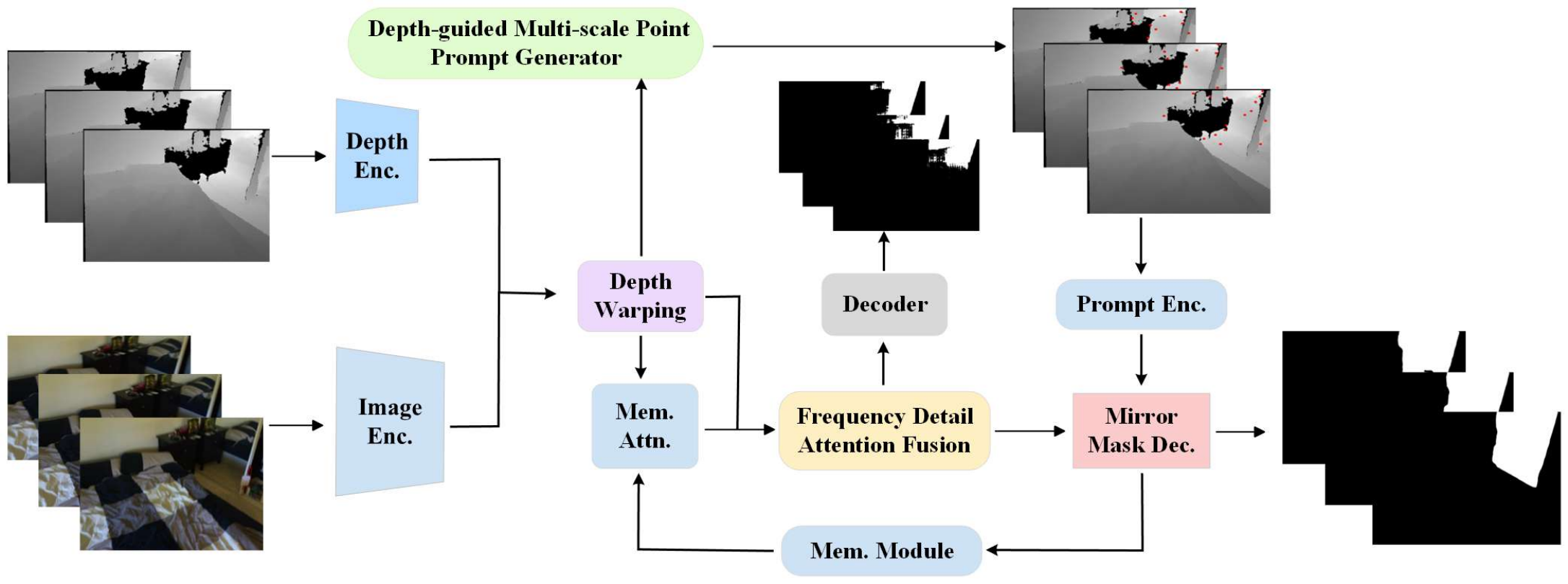} 
\caption{Overview of our MirrorSAM2 framework. Given three RGB-D video frames, the backbone extracts multi-scale RGB and depth features, while the DW module aligns depth features with RGB through cross-modal correlation. The DMS-PPG then produces spatial prompts from fused depth features. The FDAF module enhances structural cues via frequency-domain attention, and the MMD integrates these cues with a learnable mirror token to generate the final mirror segmentation mask. }
\label{method1}
\end{figure*}

\section{Related Work}
\subsection{Video Mirror Segmentation}
Video mirror segmentation has recently gained attention \cite{Lin2023, warren2024effective, xu2024zoom}.
VMDNet \cite{Lin2023} introduced the first video mirror detection network by modeling spatial-temporal correspondences.
\cite{warren2024effective} leveraged motion inconsistency, while \cite{xu2024zoom} focused on temporal variation in similarity and contrast across frames.
Although these methods improve performance, they still face challenges in handling small mirrors, ambiguous textures, and strong reflections.
Recently, \cite{Xu_2025_WACV} proposed the first RGB-D video mirror detection model and dataset, showing that depth cues can significantly enhance segmentation accuracy.

In parallel, Segment Anything Model 2 (SAM2) has shown strong generalization across segmentation tasks due to its large-scale pretraining. However, its potential for video mirror detection—particularly in automatic, prompt-free settings—remains unexplored.

\section{Method} 
\subsection{Overall Architecture}
Fig.~\ref{method1} shows the overall architecture of MirrorSAM2, which extends SAM2 for RGB-D video mirror detection. The framework consists of SAM2 and four key modules: the DW Module, the DMS-PPG, the FDAF Module, and MMD. MirrorSAM2 introduces two main improvements over SAM2: (1) It generates depth-guided spatial prompts, enhancing localization of mirror regions; (2) It refines segmentation via frequency-aware enhancement and mirror-specific decoding. We adopt a lightweight temporal sampling strategy—two adjacent RGB-D frames and one distant frame—following \cite{Lin2023, Xu_2025_WACV}, to capture both short- and long-range temporal cues while reducing computation compared to SAM2’s eight-frame input. The DW module performs cross-modal correlation to align RGB and depth features. The DMS-PPG then generates diverse, high-confidence point prompts from depth cues, which are fed into the SAM2 prompt encoder. Guided by these prompts, SAM2 predicts an initial segmentation. To address missed or ambiguous regions, the FDAF module transforms low-level depth features into the frequency domain, enriching memory tokens with fine-grained boundary information. Finally, the MMD fuses enhanced features with a learnable mirror token to produce refined and accurate mirror masks. Together, these modules enable MirrorSAM2 to effectively leverage RGB-D cues for robust, prompt-free mirror segmentation in videos.

\subsection{Depth Warping (DW) Module}
Inspired by prior works in optical flow estimation and warping-based feature alignment \cite{9093274, tian2020tdan, xie2024moving}, we design a lightweight DW module for efficient RGB–depth fusion. Unlike previous works \cite{Xu_2025_WACV} that employ a two-stream architecture to independently process image and depth features, leading to a high computational cost, we opt for a single-stream correlation-guided fusion strategy. This design choice is particularly important given the large computational footprint of SAM2. 

Given the low-level and high-level features from both the image branch \( \mathbf{F}_{\text{img}} \in \mathbb{R}^{B\times C \times H \times W} \) and the depth branch \( \mathbf{F}_{\text{depth}} \in \mathbb{R}^{B \times C' \times H \times W} \),  the DWM performs efficient fusion via correlation-guided attention and depth-aware decoding. By selectively leveraging only the low-level and high-level features, the module achieves a balance between structural detail and semantic richness while maintaining computational efficiency. The design of our DW module is inspired by the Cross-Modality Consistency Block proposed in \cite{Xu_2025_WACV}, which aligns RGB and depth features via explicit consistency modeling. In contrast, we adopt a more lightweight approach based on implicit alignment through cross-correlation and depth-guided decoding, which significantly reduces computational overhead while preserving fusion effectiveness.In addition to enhancing RGB–depth feature alignment, our DW module outputs depth features that are implicitly refined by image semantics. These refined features not only capture structural consistency but also enable more accurate and robust point generation in the DMS-PPG. The overall pipeline consists of three main steps:

First, we normalize and spatially downsample the image $\mathbf{F}_{\text{img}}$ and depth feature $\mathbf{F}_{\text{depth}}$ as $\hat{\mathbf{F}}_{\text{img}}$ and $\hat{\mathbf{F}}_{\text{depth}}$, then
compute a bi-directional correlation map using a sliding kernel-based convolution:
\begin{equation}
\mathbf{C}_{\text{img}} = \text{Corr}(\mathbf{F}_{\text{img}}, \hat{\mathbf{F}}_{\text{img}}), \quad
\mathbf{C}_{\text{depth}} = \text{Corr}(\mathbf{F}_{\text{depth}}, \hat{\mathbf{F}}_{\text{depth}})
\label{eq:dw_corr}
\end{equation}

\begin{equation}
\mathbf{C} = \frac{1}{2} \left( \mathbf{C}_{\text{img}} + \mathbf{C}_{\text{depth}} \right)
\label{eq:dw_corr_avg}
\end{equation}

This correlation is then compressed and fused back into the original features:
\begin{equation}
\tilde{\mathbf{F}}_{\text{img}} = \text{Fuse}_{\text{img}} \left(
[\mathbf{F}_{\text{img}}, \phi(\mathbf{C})] \right)
\label{eq:dw_fuse_img}
\end{equation}

\begin{equation}
\tilde{\mathbf{F}}_{\text{depth}} = \text{Fuse}_{\text{depth}} \left(
[\mathbf{F}_{\text{depth}}, \phi(\mathbf{C})] \right)
\label{eq:dw_fuse_depth}
\end{equation}
where $\phi$ is correlation crompression block ($1\times1$ conv + $3\times3$ conv), Fuse is a channel fusion layer ($1\times1$ conv + norm + RelU).
Finally, we apply a PAC-guided decoder to obtain the fused output:
\begin{equation}
\mathbf{F}_{\text{fused}} = \text{Decoder}(\tilde{\mathbf{F}}_{\text{img}}, \tilde{\mathbf{F}}_{\text{depth}})
\label{eq:dw_decoder}
\end{equation}

The output includes both the refined depth feature \( \tilde{\mathbf{F}}_{\text{depth}} \) 
and the structure-guided fusion feature \( \mathbf{F}_{\text{fused}} \), which together support 
robust point prompt generation and structural reasoning.

\subsection{Depth-Guided Multi-Scale Point Prompt Generator (DMS-PPG)}

\begin{figure}[t]
\centering
\includegraphics[width=1\columnwidth]{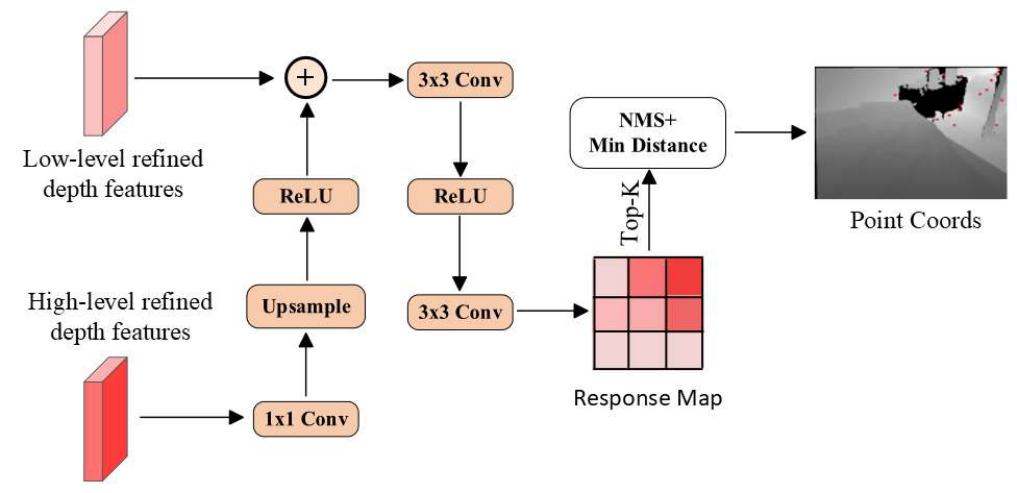} 
\caption{Illustration of our proposed DMS-PPG module. The module takes low-level and high-level refined depth features as input and fuses them in a cascade fusion way. Then generates a response map, from which top-k candidate points are selected. A smart point selection process, consisting of non-maximum suppression (NMS) and a minimum distance filter, ensures spatial diversity and semantic reliability of the final point prompts.}
\label{method2}
\end{figure}

For prompt-based segmentation models like SAM2, performance heavily relies on informative input prompts. As demonstrated in \cite{jie2024sam2meetsvideoshadow}, the absence of prompts or the use of naïve point prompts leads to significant performance degradation. To address this limitation in the context of RGB-D video mirror detection, we propose the Depth-Guided Multi-Scale Point Prompt Generator (DMS-PPG) — a key component that generates meaningful, structure-aware prompts directly from fused depth and visual cues, enabling SAM2 to function effectively without any manual prompts.

Specifically, our module generates sparse yet highly informative foreground prompts from depth-augmented multi-scale features without any manual annotation or predefined labels. These point prompts serve as soft supervisory signals that guide SAM2 to attend to object regions even in the presence of complex reflections, false symmetries, and uncertain boundaries. By embedding these points into SAM2’s transformer decoder as sparse tokens, we convert the automatic detection problem into a prompt-driven segmentation problem, solvable within the SAM2 framework.

As shown in Fig.~\ref{method2}, the DMS-PPG module takes low-level and high-level refined depth features as input. These are fused via a cascade fusion strategy: the high-level features are first projected using a 
$1\times1$ convolution and then upsampled to match the spatial size of the low-level features. The fused output is refined by a $3\times3$ convolutional block to generate a coarse response map. From this map, we apply a smart point selection process. First, the top-512 high-response candidates are selected. Then, a non-maximum suppression (NMS)-like distance filtering \cite{qi2017pointnet++} is applied to enforce spatial diversity and prevent point clustering. If no valid candidates remain, a fallback center point is used for stability. The final point coordinates are normalized to the [0,1] range and passed to the SAM2 Prompt Encoder for embedding.

Let $R \in \mathbb{R}^{H \times W}$ be the response map, then:
\begin{equation}
\resizebox{\hsize}{!}{
$\mathcal{P}_{\text{top}} = \operatorname{TopK}_N(R), \quad
\mathcal{P}_{\text{nms}} = \operatorname{NMS}(\mathcal{P}_{\text{top}}, \delta), \quad
\mathbf{x}_i = \left( \frac{x_i}{W}, \frac{y_i}{H} \right)$
}
\end{equation}
where $\mathcal{P}_{\text{top}}$ is Top-$N$ points with highest response values. $\mathcal{P}_{\text{nms}}$ is a spatially filtered set of points via non-maximum suppression (NMS). $\delta$ is the minimum distance threshold between selected points. $\mathbf{x}_i = \left( \frac{x_i}{W}, \frac{y_i}{H} \right)$ is normalized 2D coordinate of point $i$ in range $[0, 1]$. $W, H$ is the width and height of the response map. $N$ is maximum number of selected point prompts per map.

\begin{figure}[t]
\centering
\includegraphics[width=1\columnwidth]{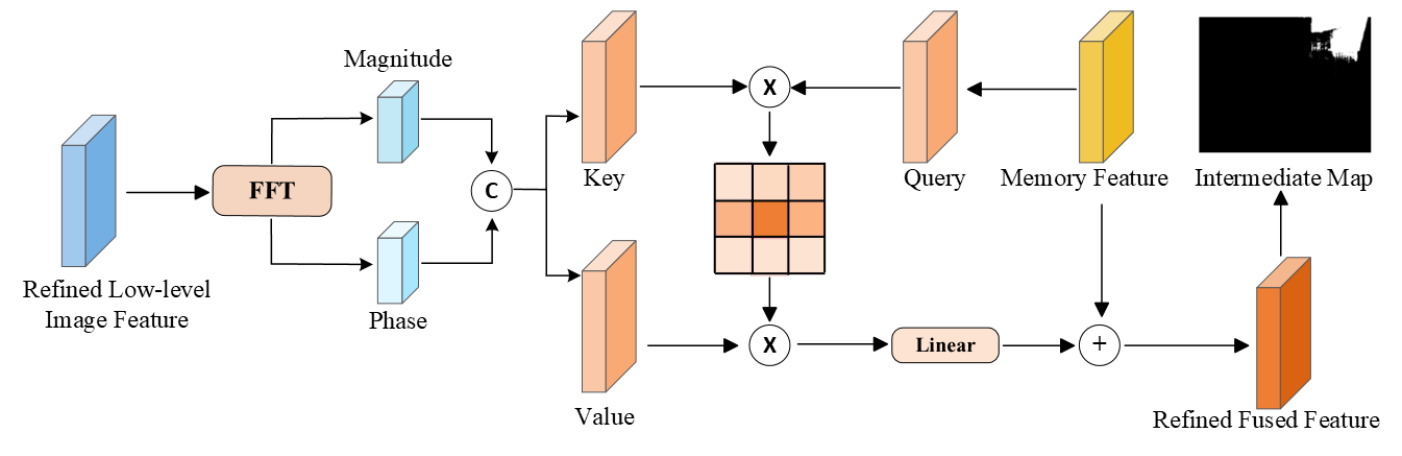} 
\caption{ Illustration of our proposed FDAF module. We use FFT to extract both magnitude and phase information. The resulting frequency features are then linearly projected to match the memory feature. Then, we treat the memory feature
as a query, and the projected frequency features as keys and values. The key and query are multiplied and forwarded to Softmax to form the attention map. Finally, the aligned frequency-enhanced output is added to the original memory features to produce the intermediate prediction map.}
\label{method3}
\end{figure}

\subsection{Frequency Detail Attention Fusion (FDAF) Module }
To enhance the detailed perception of memory features, we propose an FDAF module. This module captures high-frequency structural information from low-level depth-warped image features in the frequency domain and aligns it with memory representations via a cross-attention mechanism. This fusion strategy enriches the semantic context with fine-grained boundary cues while preserving global consistency.

As illustrated in Fig.\ref{method3}, given the structure-guided low-level fusion feature $ \mathbf{F}_{\text{fused}}^{low} \in \mathbb{R}^{B \times N_1 \times C}$, we first perform a 1D Fast Fourier Transform (FFT) \cite{cooley1965algorithm} along the channel dimension to extract both magnitude and phase information. We use both magnitude and phase from the Fourier-transformed features to capture complementary information: magnitude reflects texture strength, while phase encodes spatial structure and alignment. Phase is particularly critical for mirror detection, as it preserves symmetry and positional cues essential for distinguishing real and reflected content. This enables more precise boundary localization in reflective and symmetric regions. We adopt FFT due to its efficiency and ability to explicitly decompose features into global magnitude and phase components. Unlike Discrete Cosine Transform (DCT) \cite{ahmed1974dct} or wavelet transforms \cite{mallat1989wavelet}, FFT preserves both structural strength and spatial alignment cues without introducing extra parameters.  The resulting frequency features are then linearly projected to match the memory feature dimension. We then treat the memory features $\mathbf{F}_{\text{mem}} \in \mathbb{R}^{B \times N_2 \times d}$ as queries, and the projected frequency features as keys and values in a multi-head cross-attention module.

Finally, the aligned frequency-enhanced output is fused with the original memory features via residual connection and a feedforward block:
\begin{equation}
\hat{\mathbf{F}}_{\text{fused}} = \mathbf{F}_{\text{mem}} + \phi\left(\text{Attention}(\mathbf{Q}, \mathbf{K}, \mathbf{V})\right)
\end{equation}
where $\phi(\cdot)$ denotes a feedforward block consisting of LayerNorm, GELU, and a linear layer.

This module allows the model to take advantage of local frequency details, such as edges, textures, and symmetric structures, after semantic memory aggregation. As a result, it significantly improves the accuracy and sharpness of segmentation masks, particularly in challenging reflective or fine-structured regions.

\subsection{Mirror Mask Decoder (MMD)}
We propose MMD, a transformer-based decoder architecture designed for accurate mirror region segmentation. It integrates semantic, geometric, and frequency-aware cues via: (1) a mirror token for explicit reflective-object modeling, (2) a context contrast pathway for multiscale feature reasoning, and (3) depth-guided sparse prompts to guide the decoding process.

Formally, the decoder receive three key feature maps: $\hat{\mathbf{F}}_{\text{fused}}$ a frequency-enhanced image feature from the FDAF module, $\hat{\mathbf{F}}_{\text{fused}}^{low}$ a structure-guided low-level fusion feature from the DW module and then performs element-wise multiplication with an intermediate frequency map $M_{\text{freq}}$, resulting in a frequency-aware representation. $T_{\text{sparse}} \in \mathbb{R}^{B \times N \times D}$ sparse prmpt embeddings generated from the DMS-PPG. These prompts serve as soft object priors highlighting depth-salient regions, and are concatenated with transformer output tokens.

\subsubsection{Mirror Token and Transformer}
To explicitly guide the segmentation of reflective surfaces, we introduce a learnable mirror token \( \mathbf{t}_{\text{mirror}} \in \mathbb{R}^{D} \) into the transformer decoder. This token captures global semantics of mirror regions and attends over the spatial feature map and prompt tokens. We extend the transformer input sequence with the mirror token as follows:
\begin{equation}
\mathbf{T}_{\text{in}} = \left[ \mathbf{t}_{\text{obj}},\ \mathbf{t}_{\text{IoU}},\ \{\mathbf{t}_{\text{mask}}^i\}_{i=1}^{N},\ \mathbf{t}_{\text{mirror}},\ \mathbf{T}_{\text{sparse}} \right],
\end{equation}
where \( \mathbf{t}_{\text{obj}} \), \( \mathbf{t}_{\text{IoU}} \), and \( \mathbf{t}_{\text{mask}}^i \) are standard object, IoU, and mask tokens from the base transformer decoder, and \( \mathbf{T}_{\text{sparse}} \) represents sparse point prompts generated by the DMS-PPG. The transformer processes these tokens along with a spatial feature map \( \mathbf{F}_{\text{src}} \in \mathbb{R}^{B \times C \times H \times W} \), yielding output token embeddings:
\begin{equation}
\mathbf{H} = \text{Transformer}(\mathbf{T}_{\text{in}},\ \mathbf{F}_{\text{src}}),
\end{equation}
where the output embedding corresponding to the mirror token is denoted as:
\begin{equation}
\mathbf{h}_{\text{mirror}} = \mathbf{H}[\text{mirror index}] \in \mathbb{R}^{D}.
\end{equation}

The mirror token acts as a class-specific semantic carrier for mirror regions. Unlike general mask tokens, it is specialized to capture visual cues related to symmetry, reflection, and contextual consistency. Its global attention and token-specific decoding facilitate robust mirror segmentation, even in cluttered or ambiguous visual environments.

\begin{figure*}[ht]
\centering
\includegraphics[width=1\textwidth]{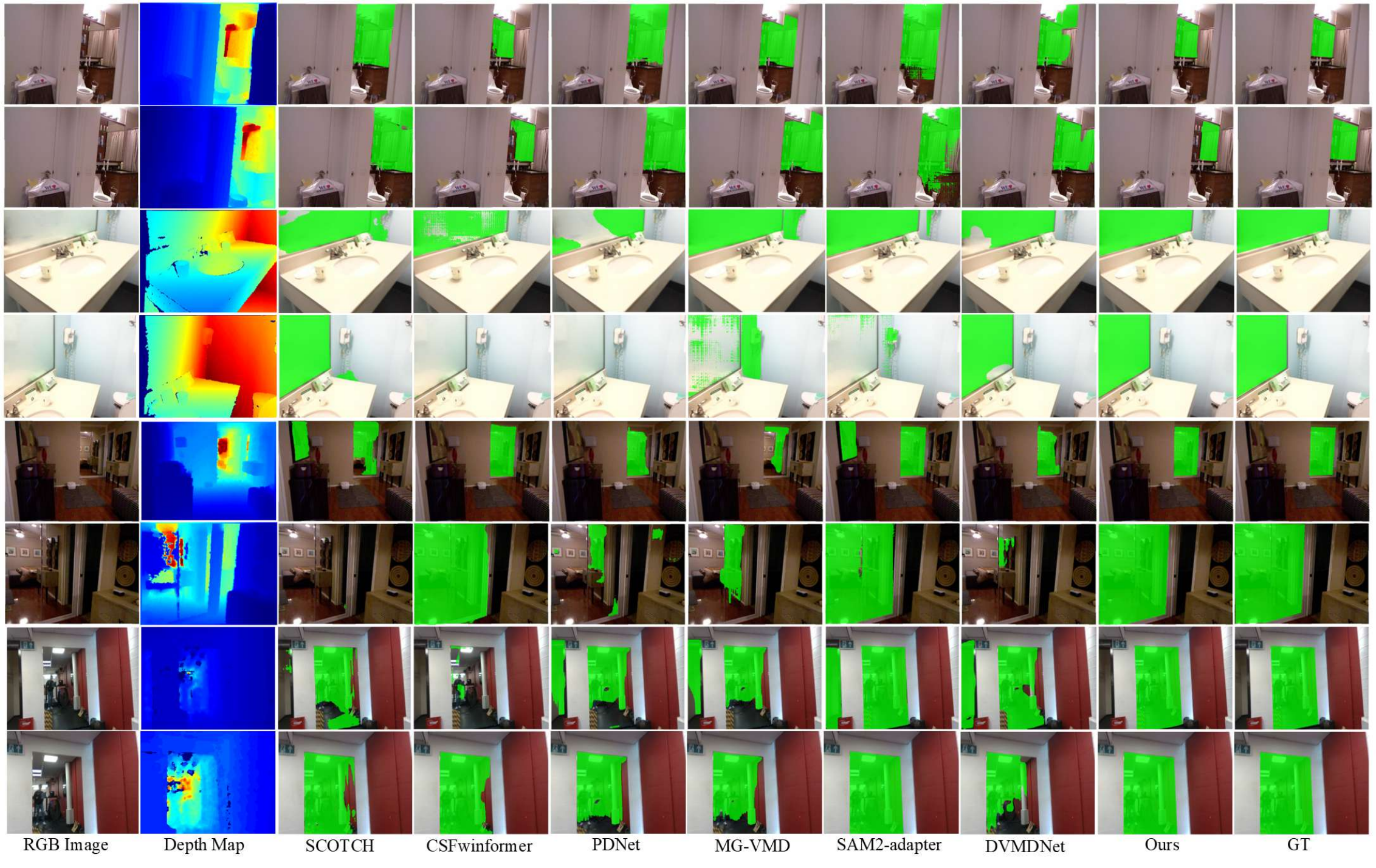} 
\caption{Visual comparison of our method with state-of-the-art methods on the DVMD dataset. }
\label{qual_res}
\end{figure*}

\subsubsection{Contrast and Fusion}
Starting from MirrorNet~\cite{Yang2019}, contrastive context modeling has proven effective in enhancing mirror detection, especially for capturing subtle boundaries and differentiating mirror from non-mirror regions. Later works such as PMDNet~\cite{Lin2020}, HetNet~\cite{he2023efficient}, SATNet~\cite{huang2023symmetry}, CSFwinformer~\cite{CSFwinformer}, ZOOM~\cite{xu2024zoom}, and DVMDNet~\cite{Xu_2025_WACV} further advanced this approach by refining contrast mechanisms for better boundary perception. Inspired by this trend, we integrate a Context Contrast Module (CCM) into our decoder, adapted from CSFwinformer~\cite{CSFwinformer}. While following its patch-wise contrast formulation, we enhance the contrast with frequency-aware and depth-aligned features.

Specifically, we extract contextual contrast features between two stages of the feature hierarchy. Given a pair of input features \( \mathbf{X}, \mathbf{Y} \in \mathbb{R}^{B \times C \times H \times W} \), we first unfold local patches via window-based extraction \( \mathcal{U}(\cdot) \), and compute their difference:
\begin{equation}
\mathbf{F}_{\text{ctx}} = \text{Attn}\left( \mathcal{U}(\mathbf{X}) - \mathcal{U}(\mathbf{Y}) \right),
\end{equation}
where \( \mathbf{X} \) is a depth-warped feature map, and \( \mathbf{Y} \) is a frequency-enhanced feature, ensuring the contrast focuses on both geometry-aligned and texture-enhanced content. Unlike CSFwinformer, which applies contrast at the output stage, our module is placed before decoding, producing intermediate contrast features at multiple levels. These features are further fused in a cascade way, and passed to a mirror token for final prediction. This reasoning module enables the decoder to capture subtle structure changes and depth-aware boundaries that are crucial for identifying mirror regions in cluttered or complex scenes.

After computing the contrastive features at two levels and fusing in a cascade way to obtain the final fused feature map $\hat{\mathbf{F}_{\text{ctx}}}$. This fused representation integrates high-level semantic structure and low-level geometric contrast. To generate the mirror mask, we use the output of the mirror token from the transformer:
\begin{equation}
\mathbf{w}_{\text{mirror}} = \text{MLP}(\mathbf{h}_{\text{mirror}}),
\end{equation}
and apply a dot product between this vector and the final mirror mask:
\begin{equation}
\mathbf{M}_{\text{mirror}}(b, h, w) = \sum_{c=1}^{C} \mathbf{w}_{\text{mirror}}[c] \cdot \hat{\mathbf{F}}_{\text{ctx}}[b, c, h, w],
\end{equation}

This design allows the mirror token to globally attend and summarize mirror-relevant information, and then decode it into a dense pixel-level prediction conditioned on depth and frequency contrast features.

\subsection{Loss Function}
We adopt a hybrid loss function $\mathcal{L}$, defined as
\begin{equation}
    \begin{aligned}
      \mathcal{L} & = \sum_{i}^{i\in\left\{t-1,t,n\right\}}\mathcal{L}_{IoU}(M_i,G_i) + \mathcal{L}_{BCE}(M_i, G_i)+ \\
      &\mathcal{L}_{IoU}(P_i,G_i)+ \mathcal{L}_{BCE}(P_i, G_i)
    \end{aligned}
    \label{eq:loss}
\end{equation}
where $\mathcal{L}_{\mathrm{BCE}}$ and $\mathcal{L}_{\mathrm{IoU}}$ denote the binary cross-entropy loss and the intersection-over-union loss, respectively, computed between the predicted mirror regions and the ground truth at the $i$-th frame. $M_i$ denotes the final mirror prediced map, $P_i$ denotes the intermediate predicted map, and $G_i$ represents the ground truth mirror map.

\begin{table*}[t]
  \centering
  \resizebox{\linewidth}{!}{
   \renewcommand{\arraystretch}{1.3} 
   \small
  \begin{tabular}{l|c|cccc|cccc}
    \hline
    \multirow{2}{*}{Methods} & \multirow{2}{*}{Pub. Year} &
    \multicolumn{4}{|c|}{VMD} & \multicolumn{4}{|c}{DVMD} \\
    && IoU$\uparrow$ & $F_\beta\uparrow$ &Accuracy$\uparrow$  &MAE$\downarrow$ &IoU$\uparrow$ & $F_\beta\uparrow$ &Accuracy$\uparrow$  &MAE$\downarrow$\\   
    \hline
    TVSD \cite{chen2021triple} & CVPR'2021 & 
    0.3397&0.5629&0.8479&0.1521 &
    0.4168&0.6841&0.8931&0.1089\\
    SCOTCH \cite{liu2023scotch}& CVPR'2023 &  
    0.5910&0.7247&0.8754&0.1236 &
    0.6223&0.7556&0.9055&0.0813\\
    \hline
    GDNet \cite{mei2020don}    & CVPR'2020 &  0.5824&0.7510&0.9206&0.0830&  0.5612&0.7339&0.9076&0.0921\\
    VGDNet \cite{liu2024multi} & AAAI'2024 &  0.5265&0.7297&0.9189&0.0821&  0.5164&0.7201&0.9068&0.0939\\
    \hline
    MirrorNet \cite{Yang2019}  & ICCV'2019 &  0.5427&0.7593&0.8709&0.1219&  0.4741&0.7010&0.9067&0.0966\\
    PMD \cite{Lin2020}         & CVPR'2020 &  0.5343&0.7895&0.8791&0.1202&  0.5210&0.7499&0.9211&0.0795\\                
    SANet \cite{guan2022learning}          & CVPR'2022 &  0.4941&0.7253&0.8729&0.1276&  0.4439&0.6214&0.9010&0.0973\\               
    HetNet \cite{he2023efficient} & AAAI'2023 &  0.5135&0.7510&0.8707&0.1216&  0.4819&0.7433&0.8991&0.1017\\  
    SATNet \cite{huang2023symmetry} & AAAI'2023 &  0.6091&0.7915&0.8901&0.0927&  0.6315&0.7931&0.9010&0.0984\\ 
    CSFwinformer \cite{CSFwinformer} & TIP'2024 &  0.6249&0.7971&0.9020&0.0905&  0.6521&0.8040&0.9029&0.0921\\
    \hline
    VMDNet \cite{Lin2023}         & CVPR'2023 & 0.5657&0.7841&0.8975&0.1029& 0.5579&0.7844&0.9051&0.0970\\
    MG-VMD \cite{warren2024effective} & CVPR'2024 & 0.6143&0.7989&0.9132&0.0945& 0.6323&0.8085&0.9239&0.0912\\
    \hline
    PDNet \cite{Mei2021}       & CVPR'2021 & 0.5621&0.7815&0.8921&0.1161  & 0.5822&0.8016&0.9204&0.0895 \\ 
    \hline
    DVMDNet \cite{Xu_2025_WACV}  & WACV'2025      & 0.6325&0.7903&0.8937&0.1062& 0.7421&0.8575&0.9492&0.0510\\
    \hline
    SAM2-adapter \cite{chen2024sam2} & ICLR'2025      & 0.7112&0.8386&0.9240&0.0759& 0.7515&0.8403&0.9480&0.0519\\
    \hline
    MirrorSAM2 & Ours      & \textbf{0.7349}&\textbf{0.8659}&\textbf{0.9273}&\textbf{0.0726}& \textbf{0.7707}&\textbf{0.8834}&\textbf{0.9544}&\textbf{0.0455}\\
    \hline
  \end{tabular}
  \renewcommand{\arraystretch}{1}
  }
  \caption{Quantitative comparison between the proposed MirrorSAM2 and 15 state-of-the-art methods from relevant fields. The best results are shown in bold.}
  \label{tab:quan_res}
\end{table*}

\section{Experiments}

\subsection{Dataset and Evaluation Metrics}
We evaluate our method in two sets of experiments: one using the RGB-D video mirror detection (DVMD)\cite{Xu_2025_WACV} dataset and the existing RGB video mirror detection (VMD) \cite{Lin2023} dataset. We use four metrics for evaluation, including intersection-over-union (IoU), F-measure ($F_{\omega}$), mean absolute error (MAE), and pixel accuracy.

\subsection{Implementation Details}
MirrorSAM2 is implemented in PyTorch, and trained using the AdamW optimizer \cite{DBLP:journals/corr/abs-1711-05101} with a learning rate of $1\times10^{-5}$ and a weight decay of $5\times 10\textsuperscript{-4}$ is utilized. All experiments and ablation studies are conducted for 30 epochs on four NVIDIA A100 GPUs (40 GB each), using a batch size of 1. Input images are uniformly resized to 1024$\times$1024 pixels using bilinear interpolation before being fed into the model.

\subsection{Comparison to SOTA Techniques}
Due to the lack of existing methods that adapt SAM2 for RGB-D video mirror detection—or even for video mirror detection in general—we refer to a few relevant works for comparison. For instance, SAM2-Adapter \cite{chen2024sam2} was proposed to extend SAM2 to downstream tasks such as camouflaged object detection, shadow segmentation, and medical image analysis. Although it is not directly designed for VMD, it provides insight into how SAM2 can be adapted for specialized domains. To date, DVMDNet \cite{Xu_2025_WACV} is the only existing method explicitly targeting RGB-D video mirror detection, and thus serves as a key point of comparison in our evaluation. In addition, we compare our approach with 13 state-of-the-art methods from related fields. These include TVSD \cite{chen2021triple} and SCOTCH \cite{liu2023scotch} for video shadow detection, GDNet \cite{mei2020don} for glass detection, VGNet \cite{liu2024multi} for video glass detection, MirrorNet \cite{Yang2019}, PMD \cite{Lin2020}, SANet \cite{guan2022learning}, HetNet \cite{he2023efficient} and SATNet \cite{huang2023symmetry} for mirror detection, VMDNet \cite{Lin2023} and MD-VMD \cite{warren2024effective} for video mirror detection, and PDNet \cite{Mei2021} for RGB-D mirror detection. All baseline models are trained and tested on the DVMD and VMD datasets using their official implementations, under the same hardware and evaluation settings. As reported in Tab.~\ref{tab:quan_res}, our method consistently achieves the best performance across all four evaluation metrics, significantly outperforming all existing state-of-the-art methods on the benchmark.
Notably, even on the VMD dataset without depth information, our method still achieves the best overall performance, demonstrating its strong generalization ability and the effectiveness of our design beyond RGB-D settings.

Fig.~\ref{qual_res} compares MirrorSAM2 with representative methods across diverse challenging scenarios. More examples are provided in the supplementary material. In the $1_{st}$ row, which involves small mirrors, only MirrorSAM2 accurately segments the mirror region on the bathroom wall. This is attributed to its integration of depth cues and frequency-enhanced structural features, enabling precise boundary detection even in cluttered textures. The $2_{nd}$ row shows a large mirror partially covered by fog, which introduces noisy depth information. Despite this, MirrorSAM2 still achieves correct segmentation by relying on its robust feature fusion, effectively combining frequency-based structure and contextual semantics. In contrast, PDNet fails due to over-reliance on depth, and CSWinFormer struggles with texture loss from the fog. The $3_{rd}$ and $4_{th}$ rows feature visually confusing non-mirror objects like paintings or doorways that resemble mirrors. These similarities mislead other methods, but MirrorSAM2 accurately distinguishes mirrors from distractors, thanks to its depth-guided point prompts and frequency-aware refinement, which provide stronger semantic discrimination.

\begin{figure}[b]
\begin{center}
  \includegraphics[width=1\linewidth]{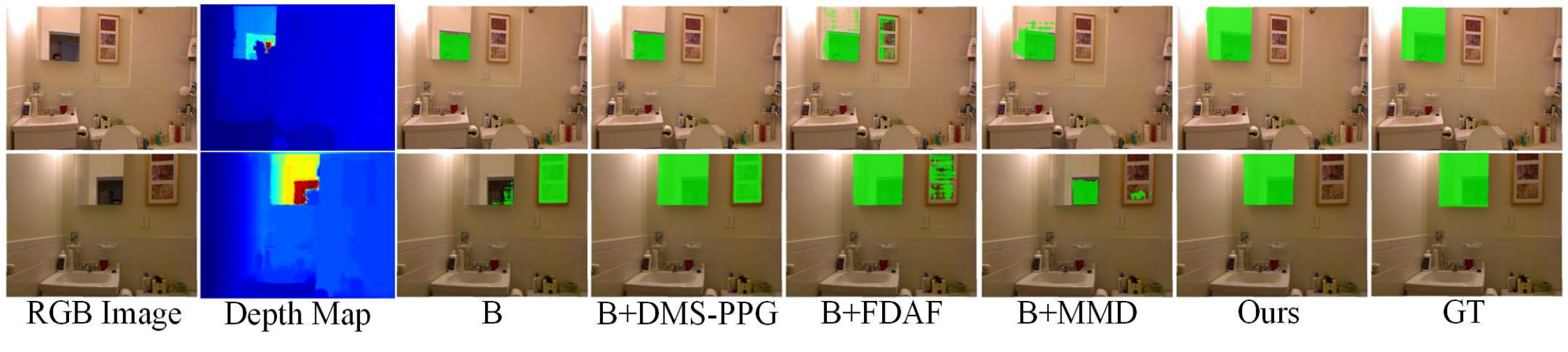}
  \caption{Visual ablation comparison of various MirrorSAM2 variants.} \label{fig:ablation_study}
\end{center}
\end{figure}

\subsection{Ablation Study}
Tab.~\ref{tab:ablation_study} illustrates the effectiveness of each component in our model. As shown in the last row, our complete MirrorSAM2, which integrates the Depth Warping MultiScale Point Prompt Generator (DWM-MPPG), the Frequency Detail Attention Fusion (FDAF), and the Mirror Mask Decoder (MMD), outperforms all other variants across all evaluation metrics.

In our ablation study, we selected SAM2 as the baseline instead of SAM2-Adapter. Preliminary experiments showed that adapter-modified features tend to lose mirror-specific cues, which are crucial for downstream modules—especially those relying on frequency representations and structural details. Using such degraded features would undermine the effectiveness of our frequency-enhancement and prompt-generation modules, making SAM2 a more appropriate and fair baseline. As shown in the $4^{th}$ column of Fig.~\ref{fig:ablation_study}, adding the DW and DMS-PPG modules significantly improves mirror localization, highlighting the role of depth-guided prompts in providing spatial priors under complex backgrounds. In the $5^{th}$ column, the inclusion of the FDAF module enhances fine structural details, demonstrating its ability to recover subtle textures and boundaries via frequency-domain information. Finally, in the $6^{th}$ column, the MMD module yields more complete and precise predictions while effectively suppressing false positives, thanks to the guidance of the learnable mirror token.

 \begin{table}[t]
 \begin{center}
    \small
    \renewcommand{\arraystretch}{1.3} 
    \begin{tabular}{l|cccc}
        \hline
        Method & IoU$\uparrow$ & $F_\beta\uparrow$ & Accuracy$\uparrow$ & MAE$\downarrow$ \\
        \hline
        B & 0.7387 & 0.8465 & 0.9388 & 0.0611 \\
        B + DW & 0.7677 & 0.8598 & 0.9491 & 0.0523 \\
        B + (DW+DMS-PPG) & 0.7691 & 0.8681 & 0.9502 & 0.0508 \\
        B + FDAF & 0.7504 & 0.8668 & 0.9516 & 0.0483 \\
        B + MMD & 0.7635 & 0.8584 & 0.9498 & 0.0501 \\
        Ours & \textbf{0.7707} & \textbf{0.8834} & \textbf{0.9544} & \textbf{0.0455}\\
        \hline
    \end{tabular}
    \renewcommand{\arraystretch}{1}

\caption{Ablation study results, trained and tested on the DVMD dataset. ``B'' denotes the original SAM2. ``DMS-PPG'' is the Depth-guided Multi-scale Point Prompt Generator. ``FDAF'' is the Frequency Detail Attention Fusion module. ``MMD'' is the Mirror Mask Decoder.} \label{tab:ablation_study}
\end{center}
\end{table}

\section{Conclusion}
In this paper, we present MirrorSAM2, the first framework to adapt Segment Anything Model 2 (SAM2) for RGB-D video mirror detection (DVMD), and also the first to explore SAM2 in a fully automatic, prompt-free setting. MirrorSAM2 incorporates four key modules: DW for RGB-depth alignment, DMS-PPG for depth-guided prompt generation, FDAF for frequency-based structural enhancement, and MMD for mirror-specific mask prediction. Experiments on VMD and DVMD-D benchmarks show that MirrorSAM2 outperforms previous methods by a large margin. While effective, challenges remain in extreme reflection or highly cluttered scenes, which we further discuss in the supplementary material.

\bibliography{aaai2026}

\end{document}